\documentclass[10pt,twocolumn,letterpaper]{article}

\usepackage{wacv}              

\usepackage{graphicx}
\usepackage{amsmath}
\usepackage{amssymb}
\usepackage{booktabs}

\usepackage{array}
\usepackage{multirow}
\usepackage{amsmath}	
\usepackage{amssymb}	
\usepackage{amscd}
\usepackage{physics}
\usepackage{rotating} 
\usepackage[toc,page]{appendix}
\usepackage[parfill]{parskip}
\usepackage{textcomp} 
\usepackage{titlesec} 

\titleformat{\section}
  {\normalfont\large\bfseries}{\thesection}{1em}{}
\titleformat{\subsection}
  {\normalfont\large\bfseries}{\thesubsection}{1em}{}
\titleformat{\subsubsection}[runin]
  {\normalfont\normalsize\bfseries}{\thesubsubsection}{1em}{}
\usepackage{changepage} 
\usepackage{setspace}
\usepackage{longtable}
\usepackage[flushleft]{threeparttable}

\usepackage[pagebackref,breaklinks,colorlinks]{hyperref}

\usepackage[normalem]{ulem}

\usepackage[capitalize]{cleveref}
\crefname{section}{Sec.}{Secs.}
\Crefname{section}{Section}{Sections}
\Crefname{table}{Table}{Tables}
\crefname{table}{Tab.}{Tabs.}

\def\wacvPaperID{232} 
\def\confName{WACV}
\def\confYear{2025}

\begin{document}

\title{SDI-Paste: Synthetic Dynamic Instance Copy-Paste for Video Instance Segmentation} 
{

\author{Sahir Shrestha
\and
Weihao Li
\and 
Gao Zhu
\and
Nick Barnes\\
The Australian National University\\
{\tt\small sahir.shrestha@anu.edu.au}
\\
{\tt\small secondauthor@i2.org}
}
\author{Sahir Shrestha \qquad Weihao Li \qquad Gao Zhu \qquad Nick Barnes \\
The Australian National University\\{\tt\small \{Sahir.Shrestha, Weihao.Li1, Gao.Zhu, Nick.Barnes\} @anu.edu.au}}
\maketitle

\begin{abstract}

Data augmentation methods such as Copy-Paste have been studied as effective ways to expand training datasets while incurring minimal costs. While such methods have been extensively implemented for image level tasks, we found no scalable implementation of Copy-Paste built specifically for video tasks. In this paper, we leverage the recent growth in video fidelity of generative models to explore effective ways of incorporating synthetically generated objects into existing video datasets to artificially expand object instance pools. We first procure synthetic video sequences featuring objects that morph dynamically with time. Our carefully devised pipeline automatically segments then copy-pastes these dynamic instances across the frames of any target background video sequence. We name our video data augmentation pipeline Synthetic Dynamic Instance Copy-Paste, and test it on the complex task of Video Instance Segmentation which combines detection, segmentation and tracking of object instances across a video sequence. Extensive experiments on the popular Youtube-VIS 2021 dataset using two separate popular networks as baselines achieve strong gains of \textbf{+2.9 AP (6.5\%)} and \textbf{+2.1 AP (4.9\%)}. We make our code and models publicly available. 

\end{abstract}

\section{Introduction}
\label{sec:intro}

\begin{figure}[!htb]
    \centering
    \includegraphics[width=\linewidth]{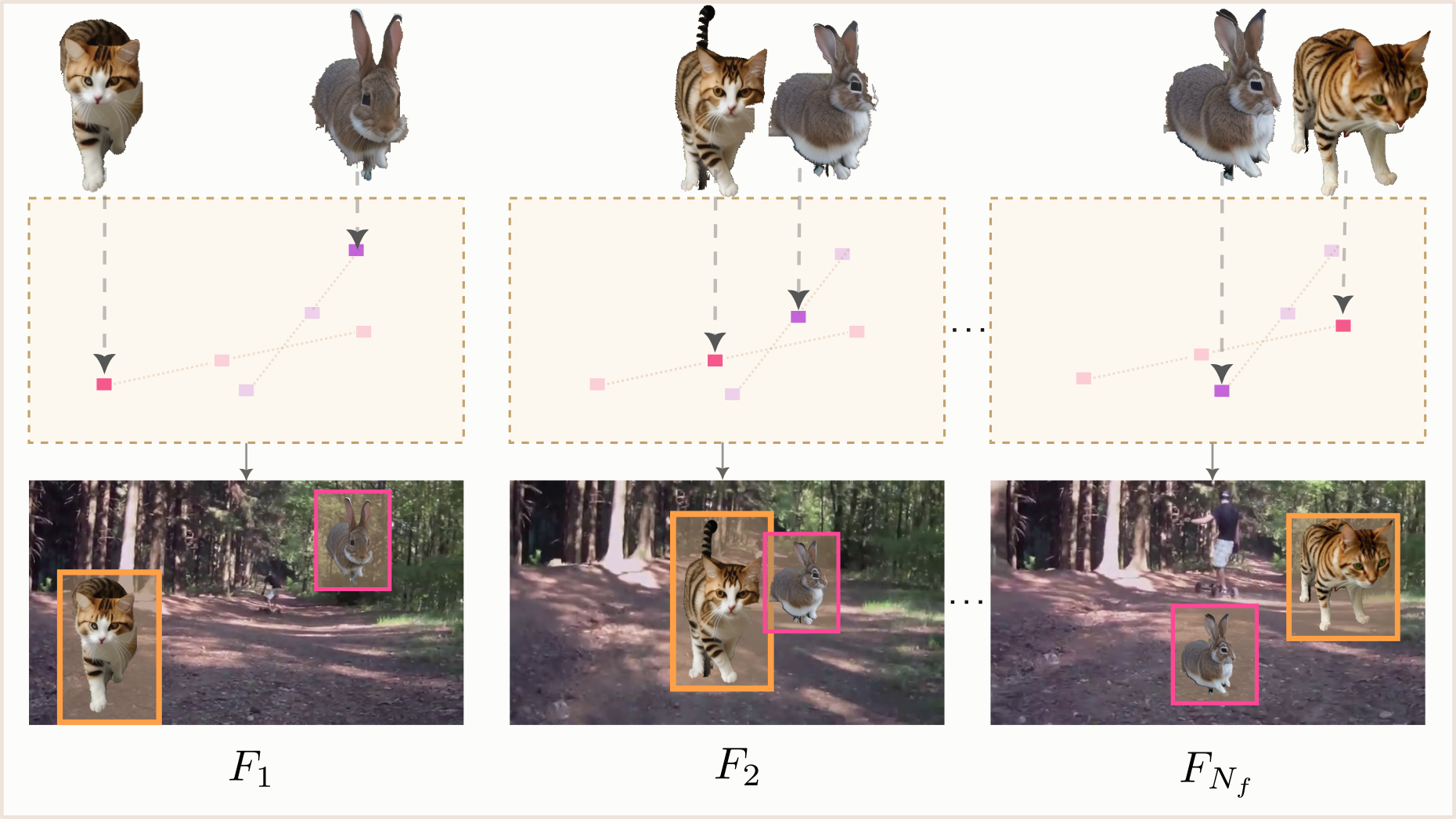}
    \caption{Our proposed data-augmentation framework generates synthetic object instances that are temporally dynamic and copy-pastes them using a linear trajectory onto each frame of a video sequence ($F_{1}, F_{2},..., F_{N_f}$). Our aim is to increase instance population of any existing video dataset. 
    } 
    \label{fig:front_page}
\vspace*{-0.5\baselineskip}
\end{figure}

Analysing video data is one of the central tasks in the field of computer vision. With the proliferation of video data today, a fundamental challenge revolves around training networks \cite{dosovitskiy2020image,he2016deep} that generalize and scale well in the face of large data diversity. It can be difficult to capture the immense variety and nuances of scenes in the real world through recorded image sequences. To tackle this, we have been relying on increasingly larger datasets \cite{deng2009imagenet, zhou2017places} to fulfill the needs of larger and deeper networks \cite{liu2021swin,radford2021learning}. However, each captured image usually requires human annotation - an endeavour that has become the central bottleneck in this pipeline as the number of recorded sequences grow. Video instance segmentation \cite{yang2019video,Ying:2023aa, wang2021end, wu2022defense, zhang2023dvis} (VIS) has emerged as a comprehensive video analysis task that encompasses recognition, segmentation and tracking of object instances across a video scene. To train a network for this task requires densely labelled image sequences where each 
object of interest is identified, labelled and its shape traced with a segmentation mask. The cost and time needed for segmentation labelling is often an order of magnitude higher than obtaining labels for other vision tasks such as classification where dense masks are not required. 
Furthermore, expanding any dataset manually requires finding suitable videos that match the complexity and scene structure of that dataset - this can prove to be a difficult affair especially for object categories that are intrinsically rare.

Data augmentations \cite{shorten2019survey} have been extensively studied as simple ways of artificially expanding a dataset. Copy-paste \cite{dwibedi2017cut,dvornik2018modeling,ghiasi2021simple,Ying:2023aa} provides an object-aware augmentation pipeline where object instances are extracted from labelled datasets using segmentation masks and pasted onto existing background images or videos. These instances are drawn from the source dataset itself \cite{dwibedi2017cut,Ying:2023aa} or from 3D models \cite{movshovitz2016useful}, neither of which provide us with a framework that is easily scalable since obtaining segmentation masks and 3D object models are both quite resource-intensive. 
Advancements in generative models have transformed computer vision in recent years.
Methods \cite{zhao2022x,fan2024divergen} have started to employ synthetic images generated from text-to-image generative models \cite{ramesh2022hierarchical,rombach2022high} to improve the performance of the copy-paste augmentation. Off-the-shelf object segmenters \cite{luddecke2022image,qin2020u2} are used to obtain new object instances from synthetic data, thereby eliminating the need for human labeling. Although synthetic data has shown to improve instance segmentation in the image domain, the application of synthetic data for copy-paste in the video domain has not yet been explored.
To this end, 
we propose a novel data augmentation method for VIS, purpose-built for dense video tasks, and enable a natural expansion of existing datasets by introducing object instances that simulate the dynamism of objects in the real world.

The recent surge in the popularity and development of diffusion models for text-to-video (T2V) generation has resulted in the creation of networks \cite{he2022latent, ho2022video, guo2023animatediff} that can generate complex video scenes, based on text prompts, with remarkable realism. Inspired by these advancements in video fidelity spearheaded by diffusion-based generative models, we explore new ways of generating, segmenting, and then incorporating synthetically generated dynamic instances of specific object categories into existing video scenes. We do this to artificially inflate the object population, aiming to synthesise a greater variety of object features in complex video scenes. We devise an infinitely scalable data augmentation framework that requires no manual dense labeling for dynamic instance generation or segmentation.




Our pipeline, using a text-to-video generative model, 
produces synthetic video scenes that capture rich object semantics through diverse viewpoints and action-states.
We extract segmentation masks for objects in each scene using an off-the-shelf self-supervised salient object segmenter. 
To ensure the validity of the generation and segmentation process, we use a zero-shot image recognition model trained on very large text-image multi-model datasets such as CLIP \cite{radford2021learning} to 
filter out any erroneously generated or segmented objects. 
Finally, we use images from YTVIS21 as background and randomly initialise the starting positions of generated objects. We explore different ways of copy-pasting dynamic objects in a video sequence and show that a linear trajectory with randomly sampled displacement gives best results. We illustrate this in Figure \ref{fig:front_page}. 

We highlight our main contributions as follows:

\begin{itemize}
  \item We propose \textbf{S}ynthetic \textbf{D}ynamic \textbf{I}nstance Copy-\textbf{Paste} (SDI-Paste) as a novel synthetic data augmentation regime for the task of Video Instance Segmentation. Our method does not require any manual dense label annotation and is infinitely scaleable.
  \item We present a pipeline for 
  on-demand crafting and segmenting of temporally dynamic objects in diverse scenes using only the category information of required objects. We then explore multiple ways of copy-pasting these instances onto existing video sequences and discover that a linear monotonic object trajectory with random jumps gives best results. 
  \item Extensive experiments 
  on the popular YTVIS21 dataset show the impressive performance of SDI-Paste. We equip two different online VIS networks with our pipeline and show a strong 6.5 \% (2.9 AP) improvement.
  We also conduct multiple ablation studies for a thorough evaluation of the different parts of our framework. We release our synthetic dataset and code-base for future video data augmentation research. 
\end{itemize}









\section{Related Work}

\textbf{Video Instance Segmentation} (VIS) \cite{yang2019video} is a dense vision task that requires joint classification, segmentation and tracking of instances across video sequences. We mainly divide the VIS methods into two groups: offline and online. Offline (or per-clip) methods \cite{lin2021video, bertasius2021classifying,wang2021end, hwang2021video} process the entire video clip simultaneously. This concurrent processing of numerous frames allows for deeper contextual understanding between them. However, these methods require significantly high processing power and memory during both training and inference. Prominent offline VIS methods either employ mask propagation \cite{lin2021video, bertasius2021classifying} or use transformer frameworks \cite{wang2021end, hwang2021video}. Online (or per-frame) methods \cite{yang2019video,huang2022minvis,yang2021crossover,wu2022defense,Ying:2023aa,heo2023generalized} carry out instance segmentation using only a small number of frames within a local range, aiming to facilitate near real-time processing. At each step, instances are detected and assembled into short sequences from the available frames.
For example, CTVIS \cite{Ying:2023aa} are built on top of image-level instance segmentation models \cite{carion2020endtoend, cheng2022maskedattention} by adding an additional pipeline that performs tracking. Memory mechanisms are utilized in IDOL \cite{wu2022defense} and CTVIS \cite{Ying:2023aa} to store instance identities as more frames become available, which improve tracking performance through consistent re-identification of object instances. While online VIS methods lag behind offline methods in terms of perfomance, they offer efficient training/inference cycle with lower memory usage. For this reason, we test SDI-Paste on online VIS. Specifically, we build and test our pipeline on CTVIS \cite{Ying:2023aa} and IDOL \cite{wu2022defense} as they are among the state-of-the-art online VIS networks.

\textbf{Video Generation.} Diffusion models \cite{sohl2015deep} have recently seen rapid development in generation of high-resolution complex image scenes and are the most popular framework for Text-to-Image (T2I) synthesis \cite{podell2023sdxl,mokady2023null}. 
Stable Diffusion \cite{rombach2022high} performs sampling in latent feature space using an autoencoding framework to enable T2I generation. More recently, T2I methods have been explored for video generation
conditioned on text prompts based on a diffusion framework \cite{ho2022video}. Methods since have improved upon this work using a pipeline where a pre-trained T2I model is used as backbone and motion/temporal modules are added and then trained using video data \cite{blattmann2023align, he2022latent, ho2022imagen, hu2022make}. 
Animatediff \cite{guo2023animatediff} uses a Stable Diffusion backbone and learns generalised motion priors to animate image scene. While it outputs video clips only 16 frames long, this is sufficient for online VIS methods as they usually take only a few input images. The generated animations can be chosen from a variety of image domain stylisations and is ready to be used out-of-the-box. All pre-trained models are made available and the code-base is well supported. For these reasons, we choose Animatediff to develop a novel data augmentation framework for VIS.

\textbf{Image-based Data Augmentations} are a low-cost, effective way to enlarge training datasets  \cite{726791,krizhevsky2012imagenet,simonyan2014very,szegedy2015going}. Copy-paste \cite{dwibedi2017cut,dvornik2018modeling,ghiasi2021simple} provides an object-aware data augmentation framework where objects are extracted from labelled datasets using segmentation masks and pasted onto existing background images. Ghiasi \textit{et al.}~\cite{ghiasi2021simple} find that random pasting of instances on a background, without blending or context, is sufficient to achieve good improvements. For augmentation pipelines such as Copy-Paste, obtaining novel instances of the right category or from varying object viewpoints/representations can be challenging due to the large volume of instances needed. There have been works that use 3D rendering to insert objects into image scenes \cite{movshovitz2016useful, karsch2011rendering, su2015render}. However, they require 3D model repositories that require human input. Recently, X-Paste \cite{zhao2022x} introduces object instances generated using Stable Diffusion \cite{rombach2022high} into the Copy-Paste framework for Instance Segmentation. Text-based image generative models \cite{rombach2022high} can produce an unlimited number of images which lends X-Paste a level of scalability that is difficult to match using existing datasets. Furthermore, the ability to generate any object provides an avenue of training networks to handle rare object classes. 

\textbf{Data augmentation strategies for Video} tasks usually involve extensions of image-based methods \cite{qian2021spatiotemporal}. While these improve performance, they fall short on leveraging the full scope of temporal dynamism that is unique to videos. DynaAugment \cite{kim2022exploring} and Group RandAugment \cite{an2022group} extend common image-based augmentation methods for video classification. Other approaches involve mixing multiple video scenes akin to the copy-paste ethos. VideoMix \cite{yun2020videomix} extends CutMix \cite{yun2019cutmix} by cutting and pasting frames from two different scenes. SV-Mix \cite{tan2023selective} and Learn2Augment \cite{gowda2022learn2augment} combine CutMix and Mixup \cite{zhang2017mixup} in a learnable framework. Zhao \textit{et al.}~\cite{zhao2024leveraging} experiment with spatial and feature based augmentations for Video Object Tracking while Lee \textit{et al.}~adapt Copy-Paste for Video Inpainting \cite{lee2019copy}. However, we found no works exploring data augmentation for VIS. 

These methods augment data by transforming or combining existing labelled videos. For augmentation pipelines such as Copy-Paste, obtaining novel instances of the right category or from varying object viewpoints/representations can be challenging as many instances are needed. Some works use 3D rendering to insert objects into image scenes \cite{movshovitz2016useful, karsch2011rendering, su2015render}. However, they employ 3D model repositories that require human input. This is costly for tasks like VIS that require many object types and unique instances. 


In \cite{zhang2019self}, Zhang \textit{et al.} use a Generative Adversarial Network to produce a ``dynamic" image by compressing the temporal information of foreground objects from videos into one single static image. The process of sampling to compress video information into a single static image is lossy - we posit valuable information is likely to be lost. Furthermore, this process is unsuitable for VIS as it requires dense segmentation masks for each frame of a video sequence whereas their ``dynamic" sampling always treats an entire video sequence as a single still image. More recently, generative models have been utilised to simulate video road scenes for the task of autonomous driving \cite{wei2024editable, gao2023magicdrive, li2019paralleleye}. While these approaches also use diffusion-based generative models for synthetic scene generation, their focus is on video scenes limited to street-view elements (such as cars, trees, building, etc) and built specifically for autonomous driving needs. 


In this work, we propose a pipeline that leverages text-to-video models within a Copy-Paste framework that generates and integrates object instances of diverse categories. This pipeline is designed to be easily adapted for any video task involving segmentation or tracking of diverse object instances. We aim to inject synthetically-generated, temporally-dynamic instances to expand the pool of object instances in existing video datasets. We investigate this pipeline's effectiveness and compare with new baselines as, to the best of our knowledge, ours is the first work to explore data augmentation strategies specifically for VIS.

\begin{figure*}[!htb]
    \centering
    \includegraphics[width=\linewidth]{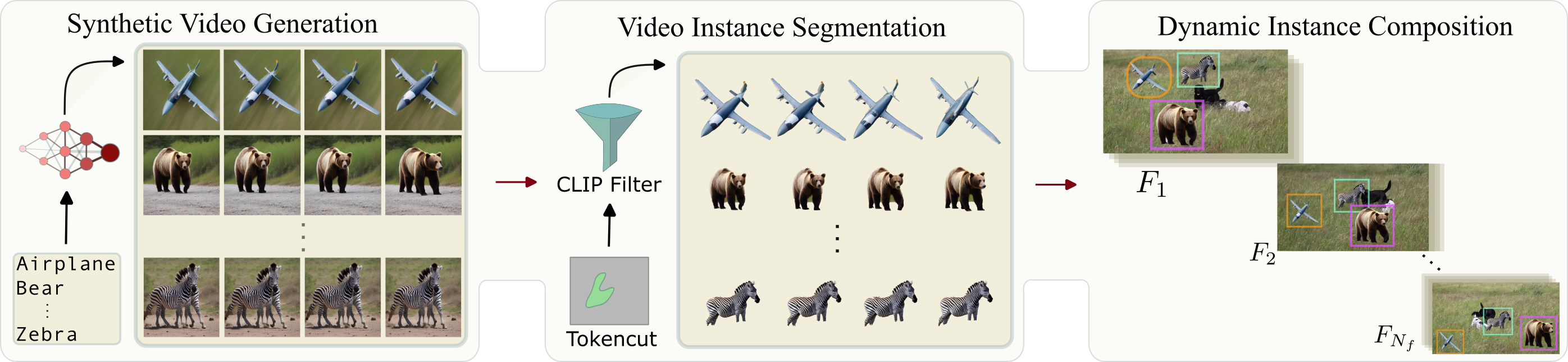}
    \caption{Illustration of our SDI-Paste pipeline. Firstly, Synthetic Video Generation uses text prompts to obtain diverse video scenes. Secondly, frames in each scene are segmented to acquire synthetic dynamic object instances. Finally, using a linear-random trajectory scheme, these dynamic instances are copy-pasted onto existing video sequences to compose the augmented dataset.}
    \label{fig:methodology}
\end{figure*}


\section{Methodology}

In this work, we aim to investigate an effective data augmentation scheme for dense video tasks by incorporating synthetically generated data. A straightforward method 
to achieve this goal is to extract object instances from static images and paste the same instance repeatedly into each frame of a video sequence. However, static instances fail to represent the inherent dynamism of real world objects that is captured in video. Without considering the dynamic nature of video data, achieving satisfactory performance is challenging. 
 
This work proposes a natural extension of the copy-paste framework, specifically designed for video, demonstrating that dynamic generative object features provide sufficient realism and supervision to improve performance. We employ a text-to-video generative diffusion model to create photo-realistic video frames. These frames are segmented and filtered using a zero-shot classification model to obtain object instance sequences. Finally, we insert these object instances as a dynamic sequence in moving locations throughout the successive frames of a video clip. 

We name our framework  \textbf{S}ynthetic \textbf{D}ynamic \textbf{I}nstance Copy-\textbf{Paste}  (SDI-Paste) and show comprehensive testing for the challenging task of VIS. Our SDI-Paste pipeline includes three main steps: Synthetic Video Generation, Video Instance Segmentation, and Dynamic Instance Composition. Details follow below.

\subsection{Synthetic Video Generation}

\def\figwidth{0.13\linewidth}
\begin{figure*} [!htb]
    \centering
    \includegraphics[width=\figwidth]{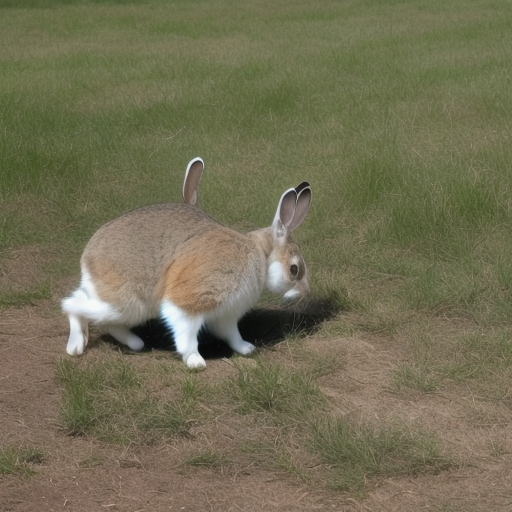}
    \includegraphics[width=\figwidth]{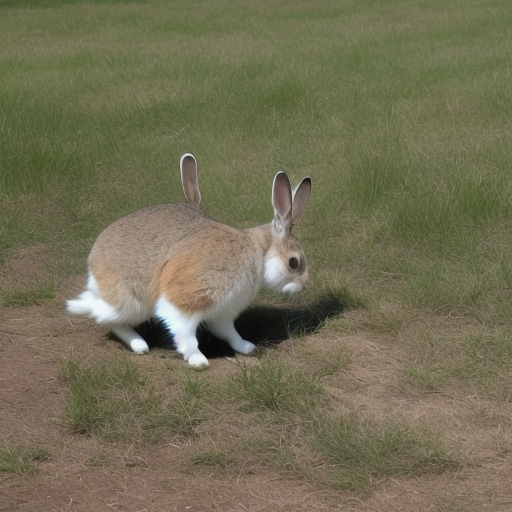}
    \includegraphics[width=\figwidth]{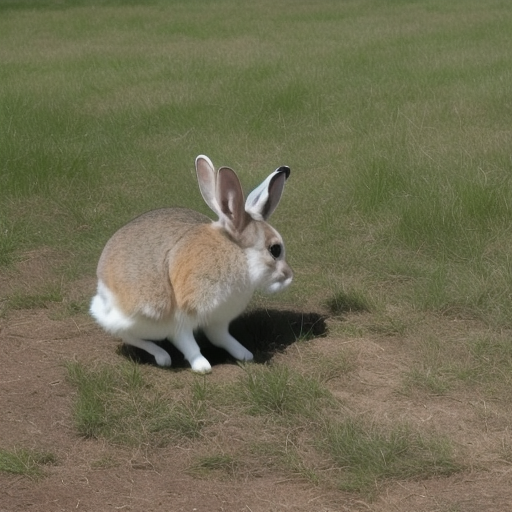}
    \includegraphics[width=\figwidth]{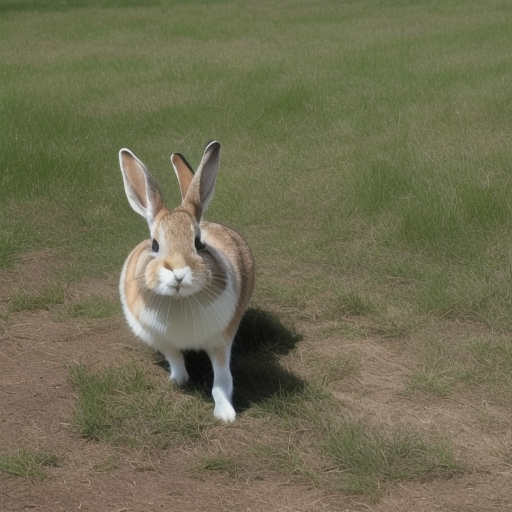}
    \includegraphics[width=\figwidth]{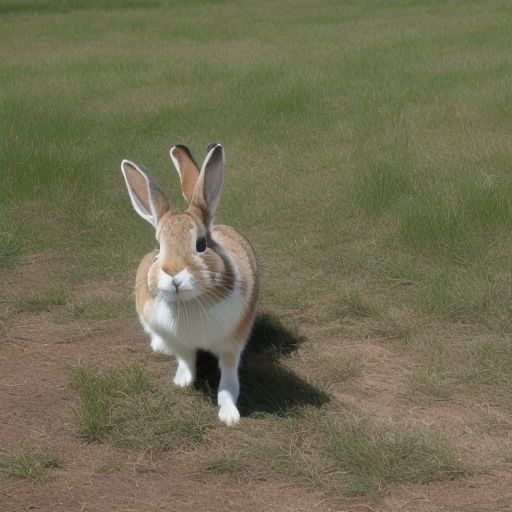}
    \includegraphics[width=\figwidth]{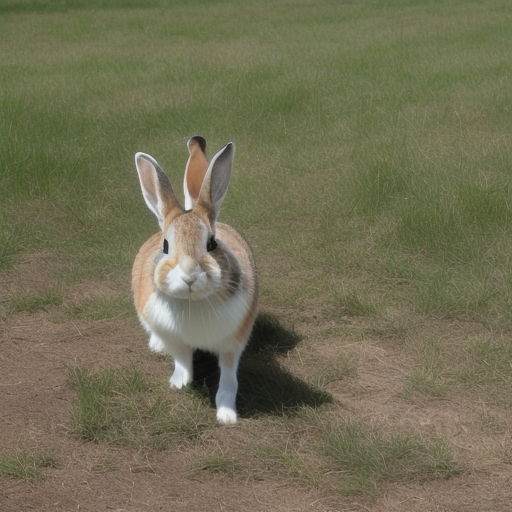}
    \\
    \includegraphics[width=\figwidth]{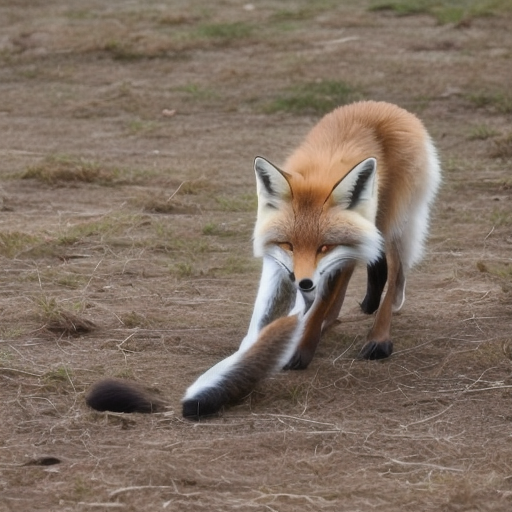}
    \includegraphics[width=\figwidth]{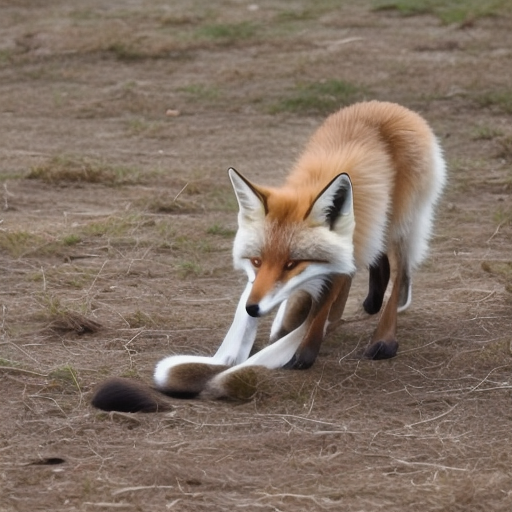}    \includegraphics[width=\figwidth]{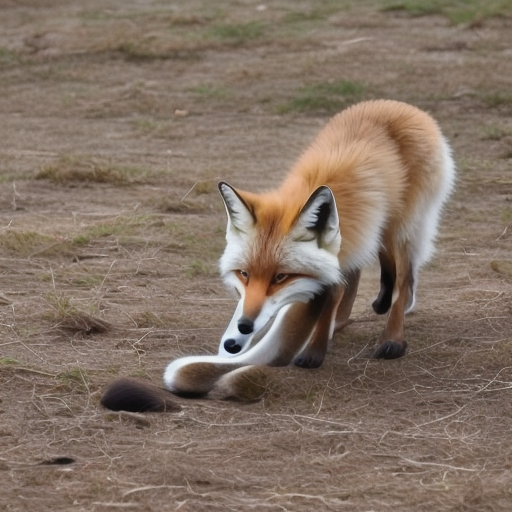}
    \includegraphics[width=\figwidth]{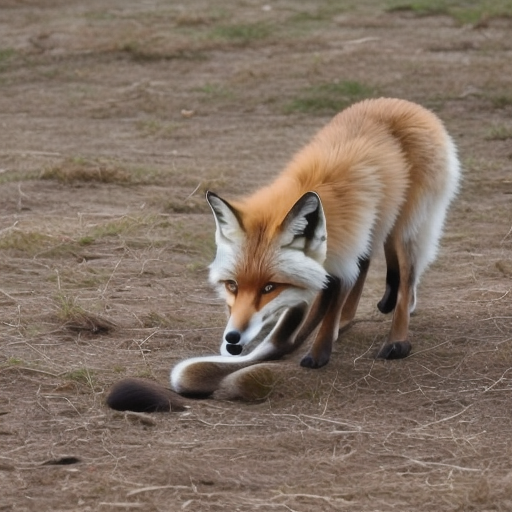}
    \includegraphics[width=\figwidth]{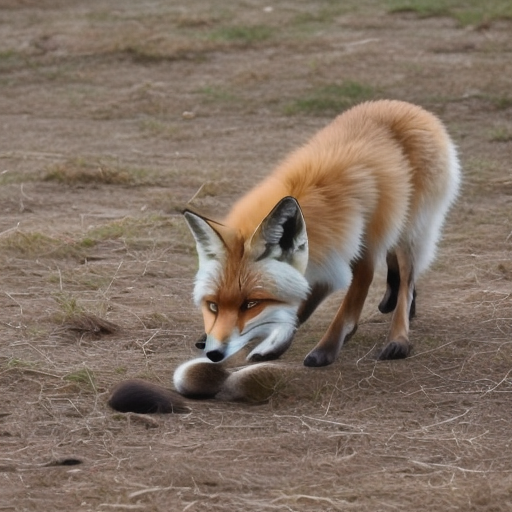}
    \includegraphics[width=\figwidth]{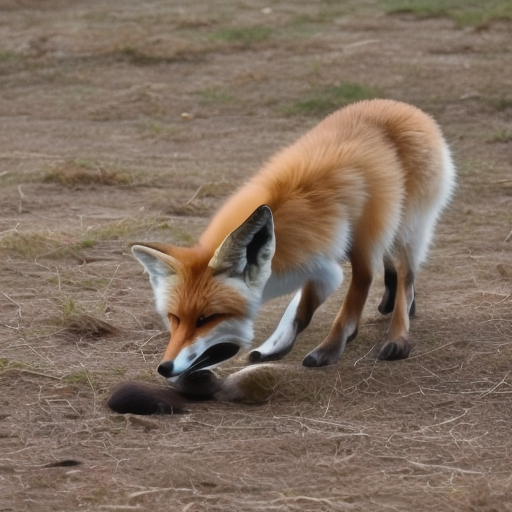}
    \\
    \includegraphics[width=\figwidth]{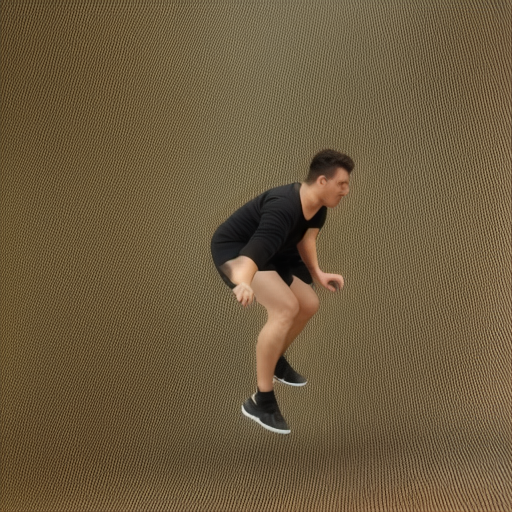}
    \includegraphics[width=\figwidth]{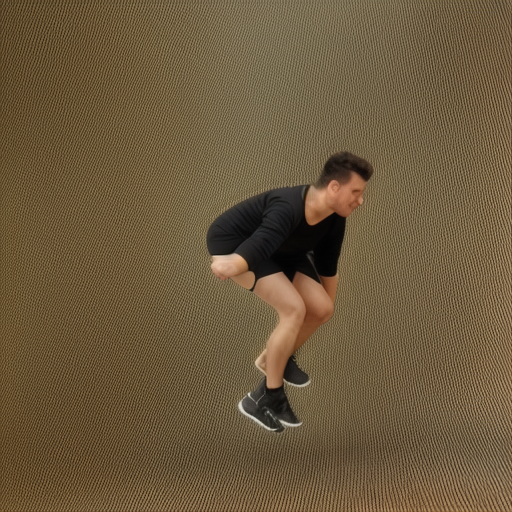}
    \includegraphics[width=\figwidth]{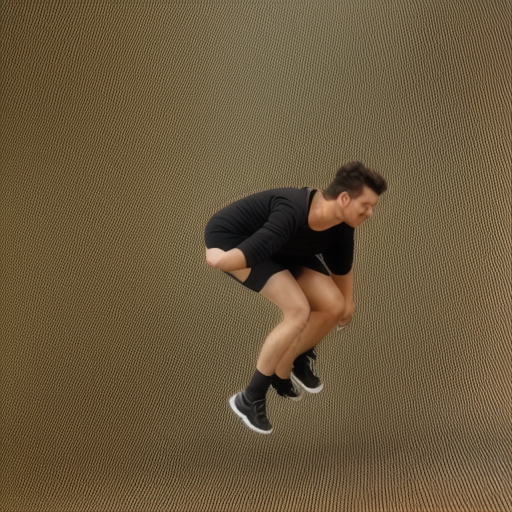}
    \includegraphics[width=\figwidth]{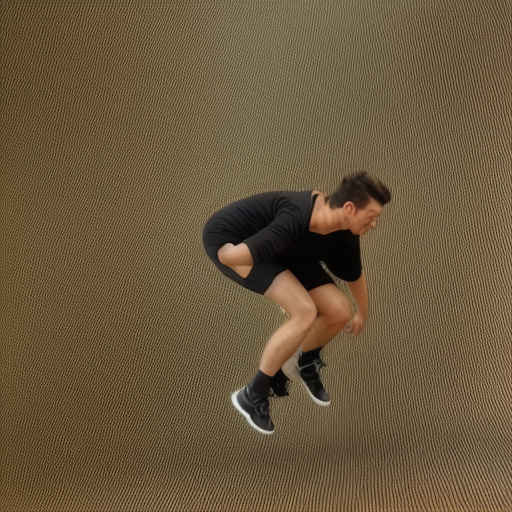}
    \includegraphics[width=\figwidth]{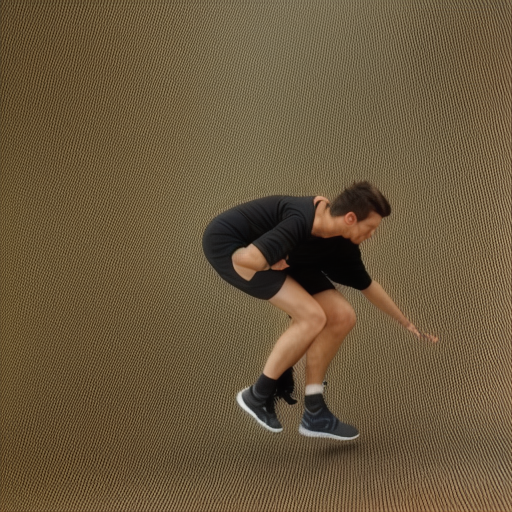}
    \includegraphics[width=\figwidth]{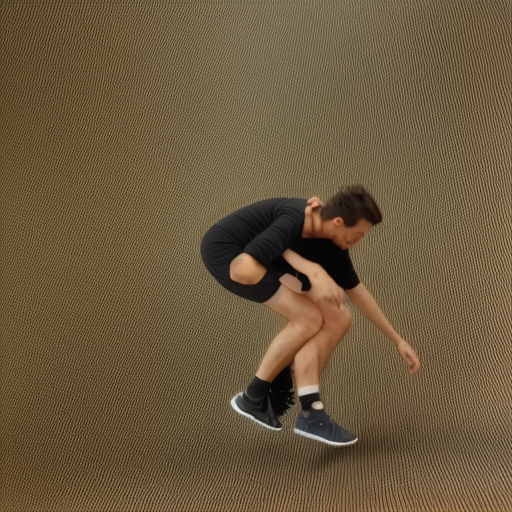}
    \\
    \includegraphics[width=\figwidth]{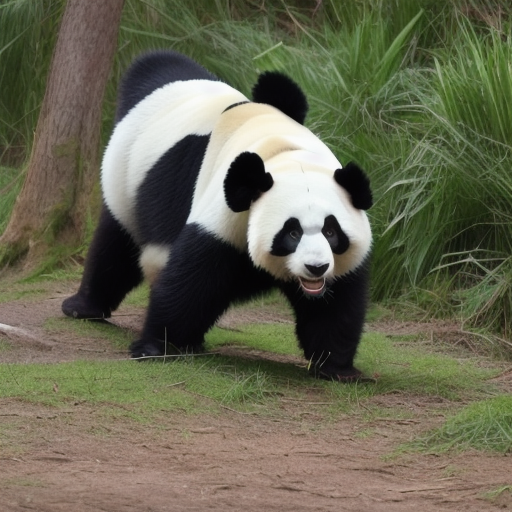}
    \includegraphics[width=\figwidth]{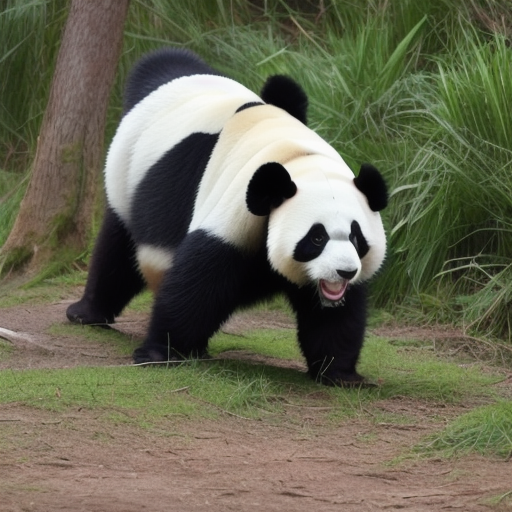}
    \includegraphics[width=\figwidth]{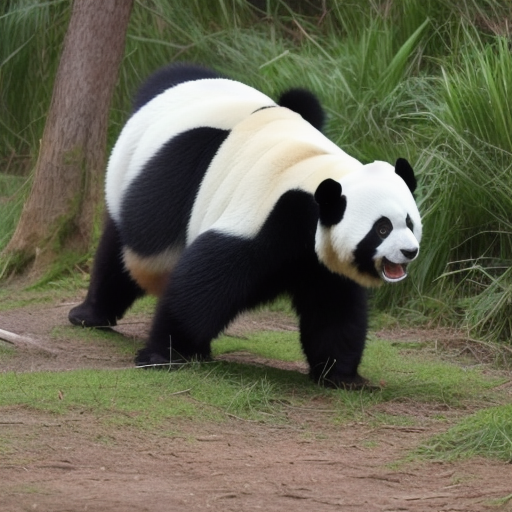}
    \includegraphics[width=\figwidth]{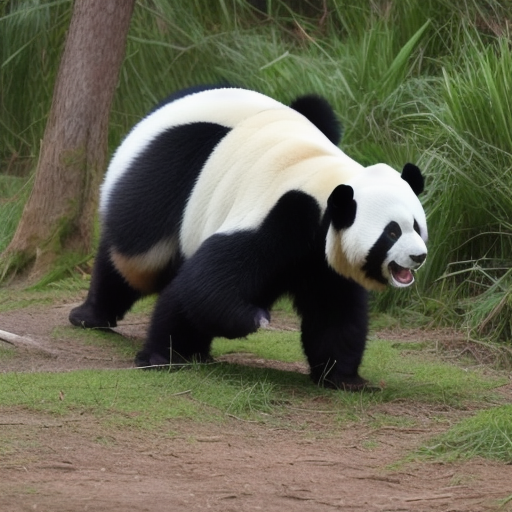}
    \includegraphics[width=\figwidth]{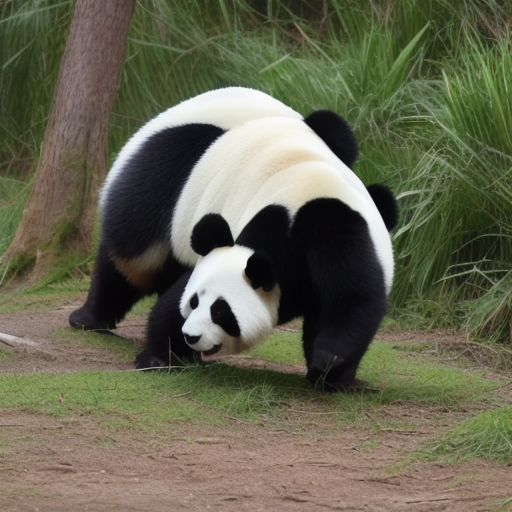}
    \includegraphics[width=\figwidth]{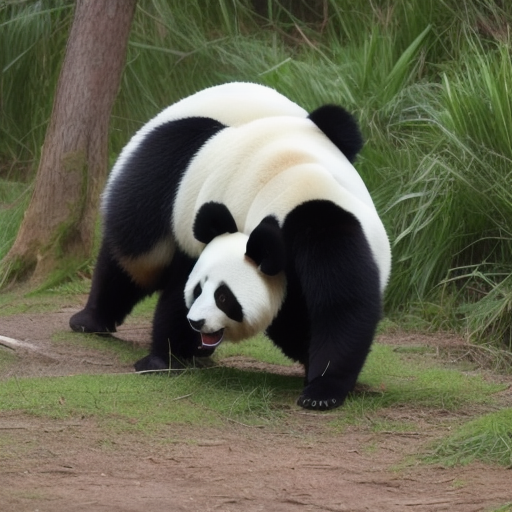}
\caption{Examples of dynamic video frames generated with AnimateDiff \cite{guo2023animatediff}. We can observe single salient foreground objects undergoing seamless shape and viewpoint transitions as a result of their actions. Object features are mostly preserved barring some aberrations such as extra ears or feet.}  
\label{fig:generated_sequences}
\end{figure*}



To generate synthetic videos, we employ a text-to-video model, such as AnimateDiff \cite{guo2023animatediff}, which incorporates motion dynamics into generated static image scenes. AnimateDiff animates images obtained through Stable Diffusion by using a separate motion modeling module trained to learn motion priors from large-scale video datasets. 
In this work, we aim to investigate augmentation using images featuring objects that change smoothly and dynamically over time, unlike Stable Diffusion, which exclusively produces static images \cite{zhao2022x}.

We generate animated sequences for specific classes by providing the text-to-video model with a sentence prompt that describes the object in a dynamic scene. Through empirical experiments, we discovered that a simple yet effective way to produce a diverse set of objects and action-states was to include generic adjectives such as ``moving" and ``dynamic" to describe the object and to place it in a ``changing" background. For example, to generate multiple and varied scenes for the object class ``bear", the text prompt will be:

\textit{
A close up video of one \textit{moving} \textit{dynamic} \textit{bear} in \textit{changing} background, moving camera, centred.}

where \textbf{bear} is a variable depending on the object class. 
We use this sentence as the input to AnimateDiff. We find that this same sentence, when run repeatedly, results in a new video scene with visually unique objects and diverse action-states each time. We use the object categories from YouTube-VIS \cite{yang2019video} to generate the necessary quantity of text inputs. When each of these text inputs is passed to AnimateDiff, it results in a short video clip comprising 16 frames. 

We show some examples of synthetic frames in Figure \ref{fig:generated_sequences} where dynamic objects naturally morph over time. We found that the generative model sometimes introduce small feature aberrations, 
such as an extra ear on the rabbit (row 1) or additional feet on the fox (row 3).
However, despite these deformed features that become visible upon close inspection, these objects are still immediately visually recognisable. 
We demonstrate that incorporating these aberrations into the augmentation process leads to a remarkable improvement in performance. We hypothesize that these aberrations provide an extra challenge to the network as it learns to not only identify the correct classifications but also learns to track as objects morph and deform due to changing features, viewpoints and actions. Furthermore, in some instances, these aberrant features can be seen to simulate the sudden appearance/disappearance of object features (such as a limb) that might manifest in a real scene.  


\subsection{Video Instance Segmentation}

The next step is video instance segmentation where we acquire segmentation masks for objects in all generated video frames. Since each generated frame has a single salient object against a generic background, we can use any off-the-shelf salient object segmentor to extract foreground instance masks. 
In our work, we use TokenCut \cite{Wang:2022aa}, a graph-based algorithm that leverages features obtained from a self-supervised transformer to detect and segment salient objects in images and videos.

To ensure the segmentation masks successfully extract the foreground object, we filter them using CLIP \cite{radford2021learning,zhao2022x}. In our setting, CLIP is employed as an assessor that matches the input text prompt and the resulting image content to obtain a relevance score for each frame. This score is used to judge the semantic relevance of the generated image and filter out instances with failed generations or erroneous segmentations. Additionally, masks occupying less than 5\% or more than 95\% of the total image area are removed.  



\subsection{Dynamic Instance Composition} \label{sec:vid_instance_composition}

\begin{figure} [!ht]
    \centering
    \includegraphics[width=0.9\linewidth]{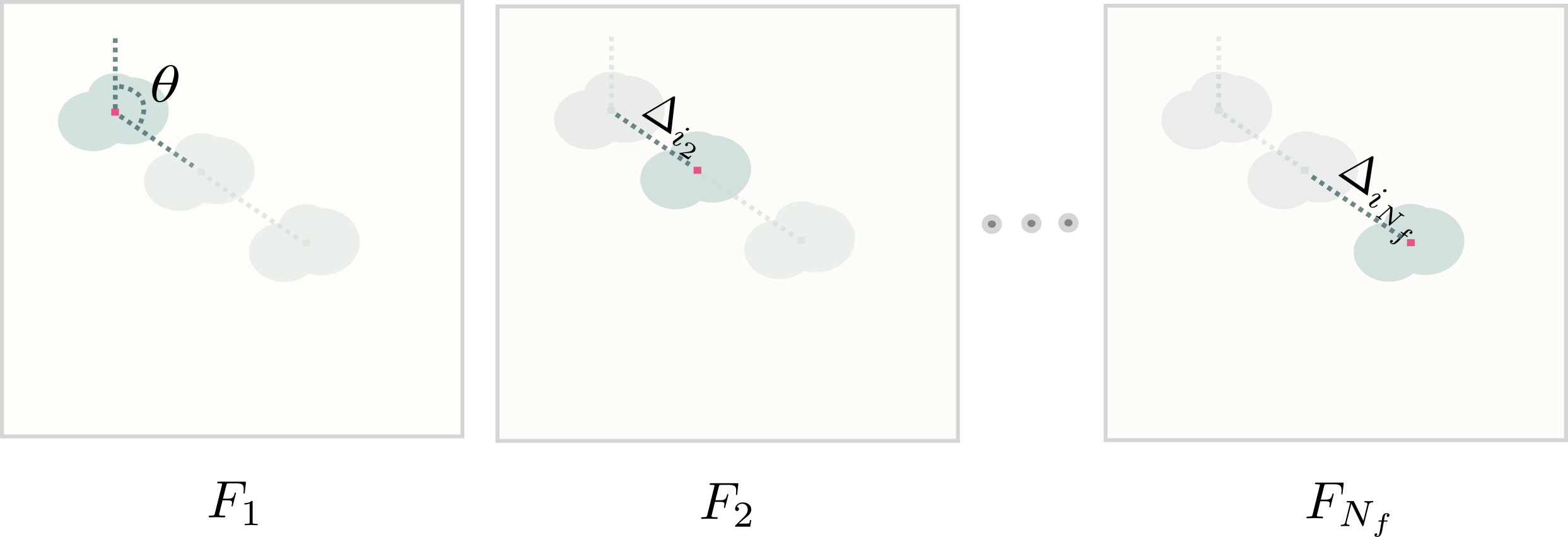}
    \caption{Figure showing linear instance placement trajectory. The direction $\theta$ is constant for all frames but the displacement $\Delta$ varies frame by frame.}
    \label{fig:instance-placement}
\vspace*{-1.0\baselineskip}
\end{figure}

\def\figwidth{0.17\linewidth}
\begin{figure*} [t]
    \centering
    \includegraphics[width=\figwidth]{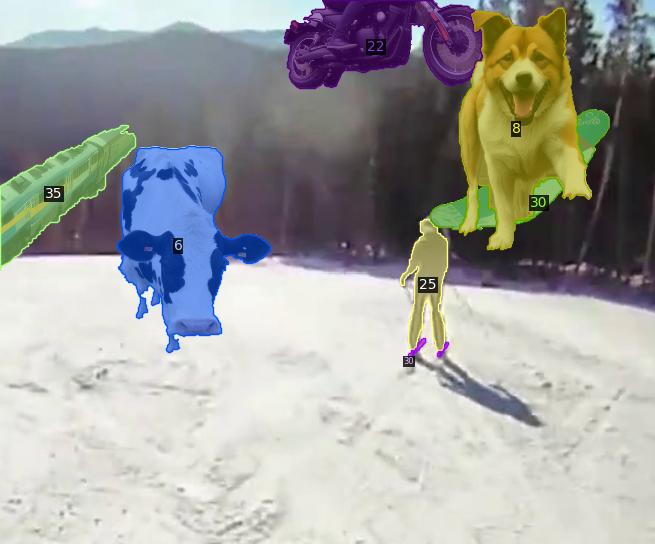}
    \includegraphics[width=\figwidth]{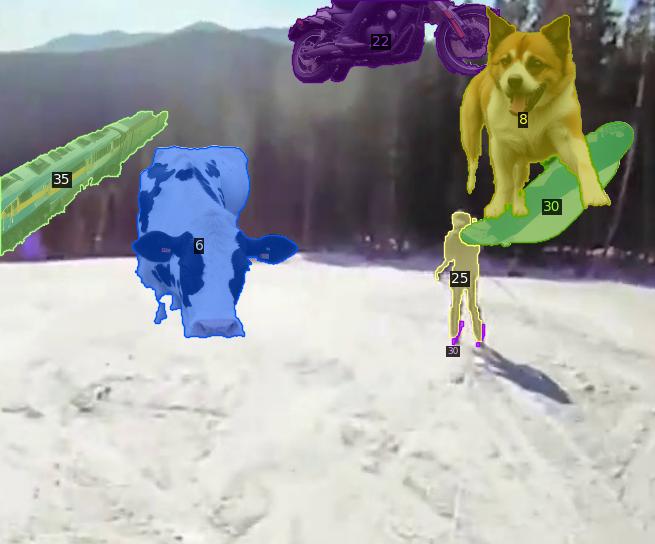}
    \includegraphics[width=\figwidth]{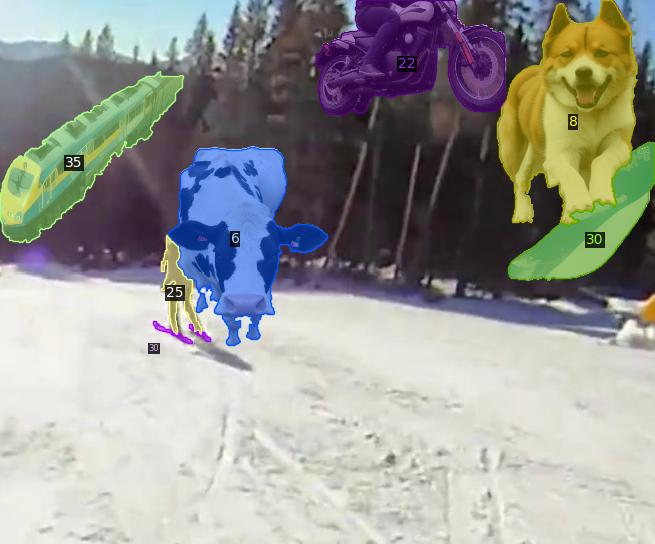}
    \includegraphics[width=\figwidth]{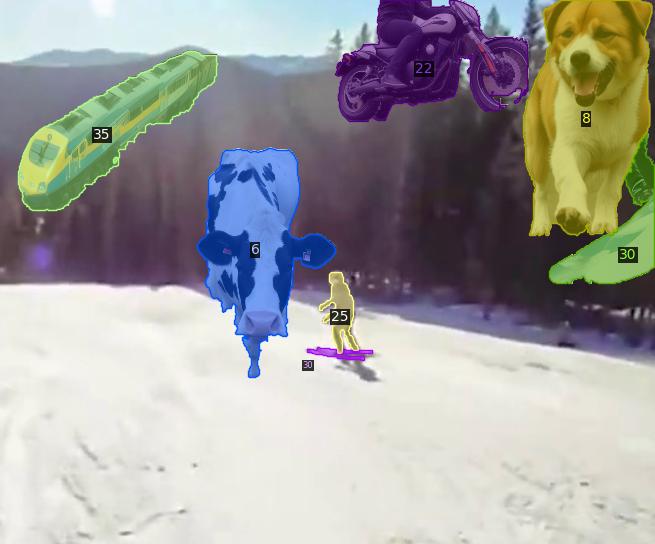}
    \includegraphics[width=\figwidth]{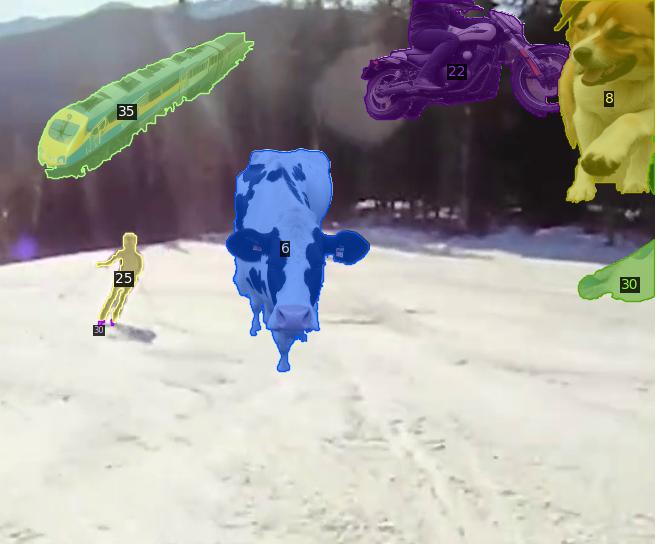}
    \\
    \vspace{0.1cm}
    \includegraphics[width=\figwidth]{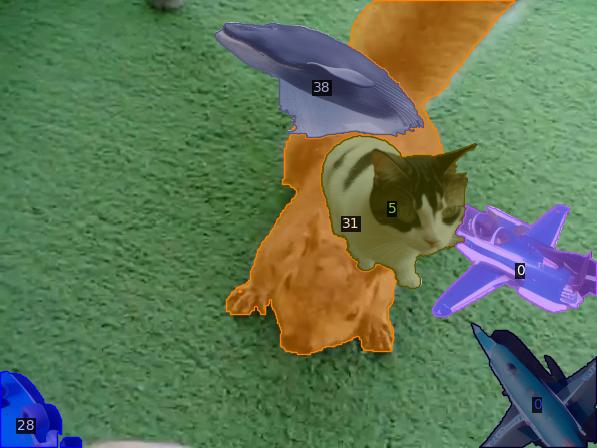}
    \includegraphics[width=\figwidth]{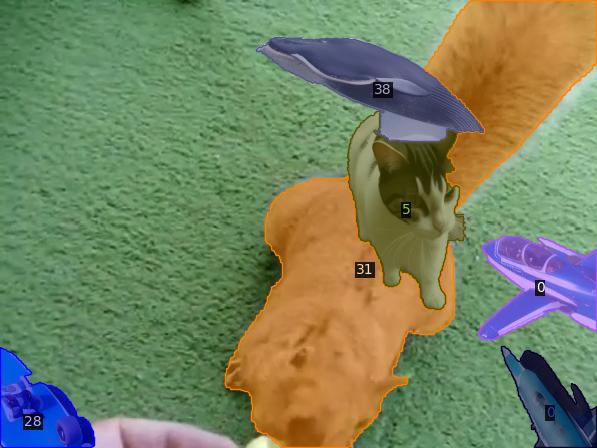}
    \includegraphics[width=\figwidth]{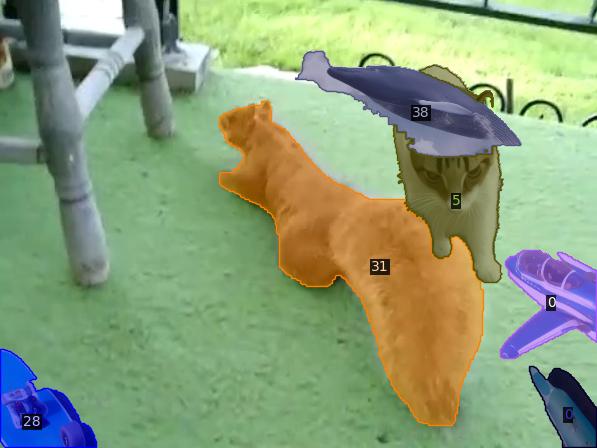}
    \includegraphics[width=\figwidth]{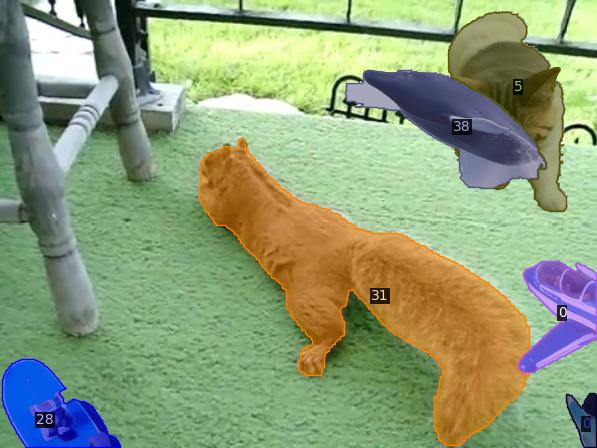}
    \includegraphics[width=\figwidth]{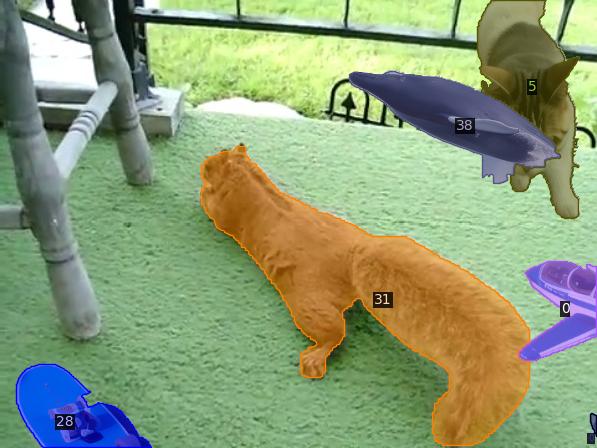}
    \\
    \vspace{0.1cm}
    \includegraphics[width=\figwidth]{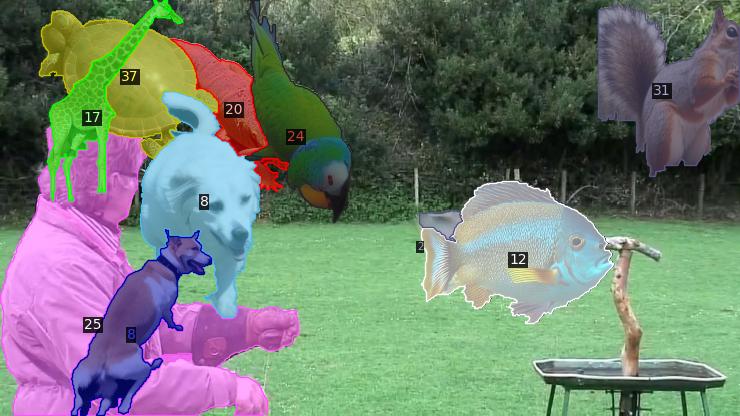}
    \includegraphics[width=\figwidth]{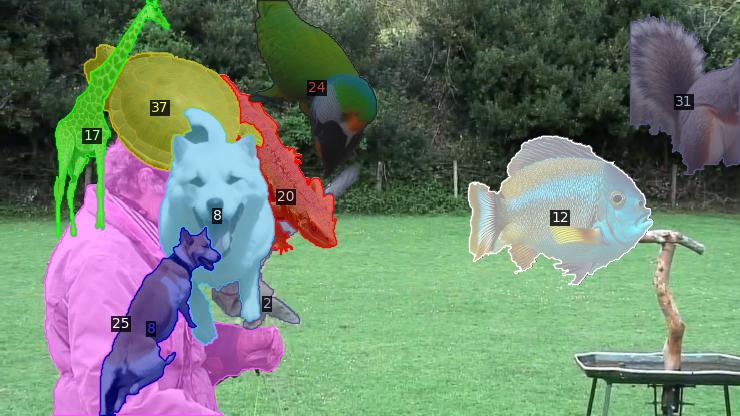}
    \includegraphics[width=\figwidth]{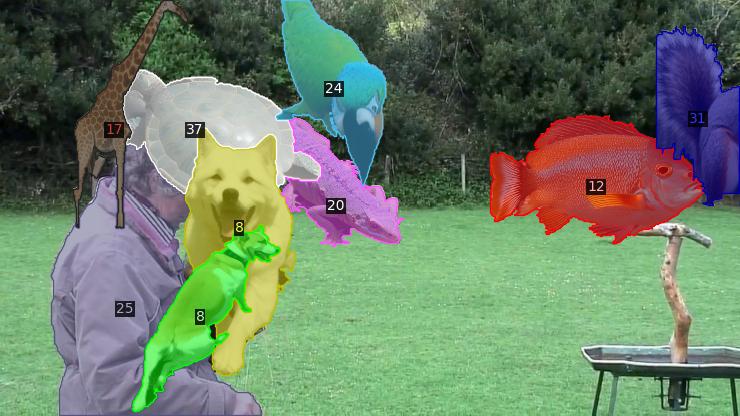}
    \includegraphics[width=\figwidth]{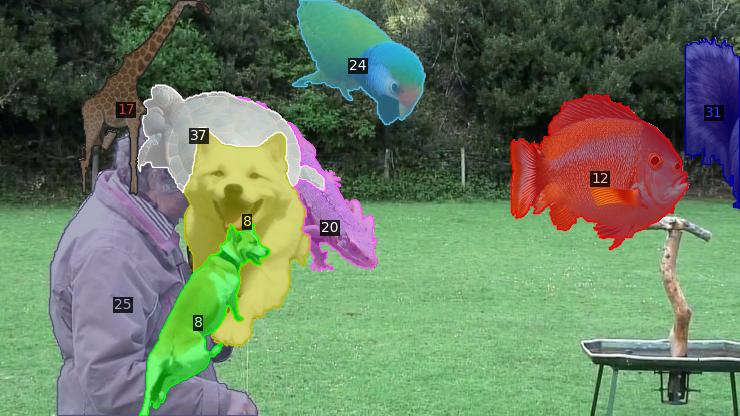}
    \includegraphics[width=\figwidth]{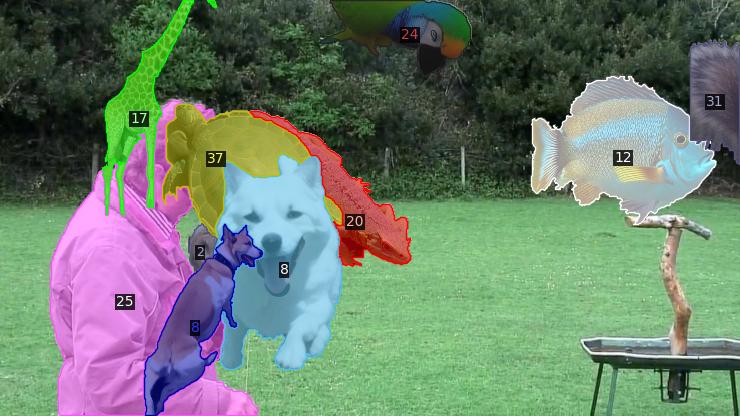}
    \\
\caption{Example of dynamic video frames obtained after instance composition. Dynamic Instances are copy-pasted onto a background image with its existing objects to enlarge the instance pool for each sequence.}
\label{fig:composed_instances}
\end{figure*}

We employ a class-balanced strategy \cite{zhao2022x} to sample instances from the segmented masks of the generated objects, and then randomly paste them onto a sequence of background frames. We use videos directly from the target dataset as the background and paste the generated instances on top of existing objects in each frame. Objects that are fully occluded after composition are removed.

We introduce a dynamic instance copy-paste strategy for video tasks that can be employed to a variable number of frames within a background sequence. We assume that there are $N_f$ consecutive frames in a background video sequence: \{$F_{1}, F_{2}, ..., F_{N_f}$ \}. First, we randomly select the number of novel objects $N_i$ we expect to introduce into this background sequence: $N_i \in [1, N_{max}], N_{max} = 20$. Then we sample $N_i$ animated instance categories. Each sampled category consists of $N_f$ instances ($i_1, i_2, ..., i_{N_f}$) with one instance for each frame of the background video sequence.
%
%

Starting with the first frame in the background sequence, we randomly sample the starting xy-coordinates $(x_0, y_0)$ for the first instance of each sampled category:
\begin{equation}
   x_0 \sim \mathcal{U}[0, W], y_0 \sim \mathcal{U}[0, H] 
\end{equation}
where $W$ and $H$ give the width and height of the background image respectively. For each subsequent frame, a trajectory system is imposed for the positioning of the remaining instances. Empirically, we achieved the best results following a linear trajectory for pasting instances across background frames. For each object placement, we fix a constant direction throughout the background sequence but allow for variable displacement between frames. 
Specifically, for each object instance $i_j, j\in[1, N_f]$, we randomly sample an angle over a uniform distribution: 
\begin{equation}
    \theta_{i_j} \sim [0,360]^\circ
\end{equation}
We follow the direction $\theta_{i_j}$ when pasting instances in all the following background frames. However, we allow the displacement of instance positions between frames to change randomly.  We show this process in Figure \ref{fig:instance-placement}. For each instance $i_j$, we randomly sample a displacement: 
\begin{equation}
    \Delta _ {i_j} \sim \mathcal{U}[0, \Delta_{max}]
\end{equation}
where $\Delta _{max}$ is the maximum allowable displacement and is a hyper-parameter we set during training.  
Based on $\theta_{i_j}$ and $\Delta_{i_j}$, we obtain the pixel displacement along x- and y- axis using standard trigonometry:
\begin{equation}
    (\delta x_{i_j}, \delta y_{i_j}) = (\Delta _{i_j} * \cos{\theta_{i_j}}, \Delta _{i_j} * \sin{\theta_{i_j}} )
\end{equation}
Then, we apply this to the initial instance position to obtain the xy-coordinates for each subsequent frame:
\begin{equation}
    (x_{i_j},y_{i_j}) = (x_{i_{j-1}} + \delta x_{i_j}, y_{i_{j-1}} + \delta y_{i_j})
\end{equation}

\begin{table*}[!ht]
\newcolumntype{P}[1]{>{\centering\arraybackslash}p{#1}}
\centering
  \begin{tabular}{P{3.3cm} | P{2.5cm} | P{1.5cm} | P{1.5cm} | P{1.5cm} | P{1.5cm}}

\hline 

\hline
    Method & $AP$ & $AP_{50}$ & $AP_{75}$ & $AR_{1}$ & $AR_{10}$ \\
\hline 
    CTVIS (Baseline) & 0.443 & 0.653 & 0.488 & 0.405 & 0.552 \\
    CTVIS (SDI-Paste) & \textbf{0.472} (+6.5\%) & \textbf{0.682} & \textbf{0.518} & \textbf{0.413} & \textbf{0.563} \\
\hline 
    IDOL (Baseline) & 0.427 & 0.643 & 0.463 & 0.396 & 0.539 \\
    IDOL (SDI-Paste) & \textbf{0.448} (+4.9\%) & \textbf{0.677} & \textbf{0.495} & \textbf{0.398} & \textbf{0.542} \\
    
\hline

\hline
\end{tabular}
\caption{Results comparing the performance of two online VIS networks with and without SDI-Paste.}\label{main_table}
\vspace*{-1.2\baselineskip}
\end{table*}

We follow X-Paste \cite{zhao2022x} in determining object scale as we paste instances onto the background frames. 
For each instance, we sample a scale $S_i$ from a Gaussian distribution $N(\mu_C, \sigma_C^2)$ and paste it with scale $S_i^2 H W$ on a background frame where $H$, $W$ denote the image height and width. For each category in the dataset, we calculate the mean $\mu_C$ and standard variance $\sigma_C ^2$ of object scales ($\sqrt{O_M/(HW)}$ within that category. Here, $O_M$ is the object mask area and $HW$ gives the total image area. In Figure \ref{fig:composed_instances}, we show some sample video frames after instance composition. 

\section{Experiments} 

\subsection{Implementation}
\subsubsection*{Datasets.} We train, test and evaluate our method on YTVIS21 \cite{yang2019video}, a popular VIS dataset. YTVIS21 is a smaller subset of the YouTube-VOS (Video Object Segmentation) dataset \cite{Xu:2018aa} from which 40 common object categories 
were retained. It consists of 2,900 videos each 3 to 6 seconds long with 4,883 unique objects. During training, we use YTVIS21 as the background images on which we paste generated objects.

\subsubsection*{Baseline Frameworks.} 
We design SDI-Paste as a plug-and-play dataset module that can be added to any online VIS training regime and reap immediate performance gains. To demonstrate, we test SDI-Paste on two popular online VIS frameworks: CTVIS \cite{Ying:2023aa} and IDOL \cite{wu2022defense}. For CTVIS, we test on a ResNet-50 \cite{he2016deep} backbone which is pre-trained on COCO \cite{Lin:2014aa}. The baseline is trained for 32000 iterations. To incorporate the SDI-Paste pipeline, we divide the training regimen into two parts: first we pre-train with SDI-Paste enabled for 16000 iterations. We use this as pretrained, disable SDI-Paste, and finetune for 16000 iterations on the base dataset only. We follow standard training settings as listed on \cite{Ying:2023aa}. Similarly, for IDOL, we test on pre-trained ResNet-50 \cite{he2016deep} backbone and follow the same training regimen to obtain the baseline and a version of the model trained with SDI-Paste. While CTVIS takes in 10 sequences as input at each training step, IDOL requires only 2. SDI-Paste can support up to 16 frames of input as we are limited by the video throughput of our generative network.   

While both of these methods offer models with larger backbones that compete with state-of-the-art in performance, they come with large compute resource overheads. Thus, we choose to test our pipeline on lighter versions of the models to demonstrate the efficacy of SDI-Paste as a pipeline that improves on any baseline regardless of the base network. We report standard metrics: $AP$, $AP_{50}$, $AP_{75}$, $AR_{1}$ and $AR_{10}$.      

\subsubsection*{SDI-Paste Settings.}
SDI-Paste is a modular framework comprising of a text-to-video generator (AnimateDiff) and an image foreground segmentor (TokenCut). For AnimateDiff \cite{guo2023animatediff}, we use the 40 object categories in YTVIS21 to produce animated video instances and obtain 470 video sequences for each category. Each sequence is 16 frames long resulting in 300800 generated image frames. We choose RealisticVision image stylisation for AnimateDiff as it produces the most natural looking videos. The remaining settings are set as recommended by the authors. 

Likewise, for TokenCut \cite{Wang:2022aa}, we use the recommended setting to segment each frame in a video sequence individually. Each segmented object is input into a CLIP model \cite{radford2021learning} and filtered. Only segmented objects with a clip score threshold of 0.21 are retained. We also remove objects that occupy less than 5\% and more than 95\% of the total image area since such objects are likely to have been mis-segmented. During dynamic instance composition, we set maximum number $N_{max}$ of novel objects introduced to each video sequence to be 20. 

\subsection{Results}

\subsubsection*{Main Results.}
In Table \ref{main_table}, we compare our SDI-Paste trained models against baseline for CTVIS and IDOL and see solid improvements of $6.5\%$ and $4.9\%$ respectively. We achieve this not by changing network structure or altering training strategies but purely through injection of synthetic dynamic instances onto the base training dataset. The discrepancy in improvement between CTVIS and IDOL can be attributed to the difference in number of images input to the network during each training step: CTVIS uses 10 sequential images whereas IDOL uses only two. This results in the network seeing more synthetic instances in CTVIS relative to IDOL. We cannot match the performance figures of our two baselines in comparison to their original papers \cite{Ying:2023aa} and \cite{wu2022defense} as we trained them from scratch in constrained training and dataset settings; our main aim was to demonstrate the efficacy of SDI-Paste regardless of the base network or its performance capability.

\begin{table*}[t]
\newcolumntype{P}[1]{>{\centering\arraybackslash}p{#1}}
\centering
  \begin{tabular}{P{3.3cm} | P{1.5cm} | P{1.5cm} | P{1.5cm} | P{1.5cm} | P{1.5cm}}

\hline 

\hline
    Method & $AP$ & $AP_{50}$ & $AP_{75}$ & $AR_{1}$ & $AR_{10}$ \\
\hline 
    Baseline & 0.443 & 0.653 & 0.488 & 0.405 & 0.552 \\
    CopyPaste \cite{ghiasi2021simple} & 0.455 & 0.659 & 0.503 & 0.404 & 0.546 \\
    X-Paste \cite{zhao2022x} & 0.468 & 0.681 & 0.505 & 0.405 & 0.559 \\
    SDI-Paste (ours) & \textbf{0.472} & \textbf{0.682} & \textbf{0.518} & \textbf{0.413} & \textbf{0.563} \\
    
\hline

\hline
\end{tabular}
\caption{Results comparing existing data augmentation methods with SDI-Paste using CTVIS with ResNet-50 as baseline.}\label{comparison_table}
\vspace*{-0.5\baselineskip}
\end{table*}

\subsubsection*{Comparison with other methods.}
In Table \ref{comparison_table}, we make comparisons against other related data augmentation methods. We use CTVIS as our base VIS framework and compare SDI-Paste with a Copy-Pastel\cite{ghiasi2021simple} baseline where we copy-paste instances from across YTVIS21. To test the effectiveness of dynamic object instances compared to static ones, we also compare against X-Paste \cite{zhao2022x} on YTVIS21. We use the code-base and recommended settings from \cite{zhao2022x} to generate and segment object instances and obtain comparable number of images to our SDI-Paste setting. We adapt X-Paste for a video task by pasting each static object instance repeatedly onto each frames in a video sequence. We observe that Copy-Paste improves the baseline CTVIS by 1.2 AP. X-Paste, with its synthetic static instances, posts an improvement of 2.5 AP over baseline whereas our SDI-Paste pipeline outperforms them both with an improvement of 2.9 AP over baseline (+0.4 AP on X-Paste and +1.7 AP on Copy-Paste). While these results show the effectiveness of the copy-paste framework in enabling a solid boost in model performance, we see that the VIS task is better served by SDI-Paste where dynamic object instances are injected onto the base dataset. We posit this is due to the dynamic instances capturing more diverse object features from varying viewpoints and shape deformations when compared to static instances as in X-Paste. Please note that we show the best results from a pool of maximum number of experiments that was possible within our computation budget.




\subsection{Ablation Study}

\subsubsection*{Ablating trajectory system.}
During Dynamic Instance Composition, we investigate three different methods of copy-pasting instances onto a sequence of images: Linear, Bezier and Linear-random. For the linear system, we paste instances across the video frames with a straight-line trajectory as directed by the angle $\theta_{i_j}$ with constant displacement of objects between the frames. The linear-random system adopts the same straight-line trajectory but allows for randomly sampled displacement of objects between frames (as discussed in \ref{sec:vid_instance_composition}). For the Bezier system, we trace the path of the object using a Bezier curve with a random length. Comparing these three systems, we find that the linear-random trajectory consistently gives the best result while the linear system is only slightly behind. The Bezier trajectory gives the worst performance. This might be due to SDI-Paste being introduced as a pre-training step where simple linear trajectories are easier for the network to track. Possibly, an only-Bezier-curve trajectory does not account for the majority of object movement in YTVIS21. A further experiment could be a system that combines diverse ways of moving objects improves results. We leave this for future works.

\begin{table}[h]
\newcolumntype{P}[1]{>{\centering\arraybackslash}p{#1}}
\centering
  \begin{tabular}{P{2.2cm} | P{0.68cm} | P{0.68cm} | P{0.68cm} | P{0.68cm} | P{0.68cm}}
\hline 

\hline
    Trajectory Method & $AP$ & $AP_{50}$ & $AP_{75}$ & $AR_{1}$ & $AR_{10}$ \\
\hline 
    Bezier curve& 0.452 & 0.678 & 0.505 & 0.405 & 0.542 \\
    Linear & 0.464 & \textbf{0.700} & 0.501 & 0.391 & 0.530\\ 
    Linear-random & \textbf{0.472} & 0.682 & \textbf{0.518} & \textbf{0.413} & \textbf{0.563} \\
\hline

\hline
\end{tabular}
\caption{Results comparing different trajectory systems adopted during Dynamic Instance Composition.}\label{table:ablation_trajectory}
\vspace*{-1.0\baselineskip}
\end{table}

\subsubsection*{Ablating segmentation method.}
In X-Paste \cite{zhao2022x}, Zhao \textit{et al.} employ a CLIP-guided selection of segmentation masks obtained from four different supervised salient object segmentation networks. We test this strategy against TokenCut \cite{Wang:2022aa} as the only segmenter and find that it consistently outperforms the X-Paste pipeline. 

\begin{table}[!h]
\newcolumntype{P}[1]{>{\centering\arraybackslash}p{#1}}
\centering
  \begin{tabular}{P{2.1cm} | P{0.7cm} | P{0.7cm} | P{0.7cm} | P{0.7cm} | P{0.7cm}}
\hline 

\hline
    Segmentation Pipeline & $AP$ & $AP_{50}$ & $AP_{75}$ & $AR_{1}$ & $AR_{10}$ \\

\hline 
    X-Paste CLIP-guided \cite{zhao2022x} & 0.462 & \textbf{0.699} & 0.501 & 0.399 & 0.545 \\ 
    TokenCut \cite{Wang:2022aa}& \textbf{0.472} & 0.682 & \textbf{0.518} & \textbf{0.413} & \textbf{0.563} \\
\hline

\hline
\end{tabular}
\caption{Comparison of two segmentation methods: X-Paste CLIP-guided strategy \cite{zhao2022x} and TokenCut \cite{Wang:2022aa}. }\label{table:ablation_segmentation}
\vspace*{-0.5\baselineskip}
\end{table}

\subsubsection*{Ablating effect of more instances.}

\begin{table}[h]
\newcolumntype{P}[1]{>{\centering\arraybackslash}p{#1}}
\centering
  \begin{tabular}{P{2.1cm} | P{0.7cm} | P{0.7cm} | P{0.7cm} | P{0.7cm} | P{0.7cm}}
\hline 

\hline 
    Number of sequences & $AP$ & $AP_{50}$ & $AP_{75}$ & $AR_{1}$ & $AR_{10}$ \\

\hline 
    150 & 0.456 & 0.674 & 0.505 & 0.408 & 0.559 \\ 
    470 & \textbf{0.472} & \textbf{0.682} & \textbf{0.518} & \textbf{0.413} & \textbf{0.563} \\

\hline

\hline

\end{tabular}
\caption{Results comparing the effect of increasing the number of generated dynamic instances on CTVIS. Each sequence consists of 16 frames.}\label{table:ablation_instances}
\vspace*{-1.0\baselineskip}
\end{table}
To verify the effect of increasing the number of synthetic instances generated on model performance, we create two datasets where, for each category, we generate either 150 or 470 dynamic instance sequences (each sequence consists of 16 image frames). These amount to 96000 and 300800 image frames generated respectively. We see in Table \ref{table:ablation_instances} that increasing the number of dynamic instances improves model performance by $3.5\%$. As expected, with a larger number of unique instances available for training, the network shows remarkable improvement on the same training regimen. Given our compute resources, using AnimateDiff to produce object instances was a significantly heavy task which limited our ability to test with larger generated datasets. We expect the model performance to benefit further from an even larger instance pool to draw from. We make the generated dataset, as well as code to reproduce it at any size, available and leave this task for future research.

\section{Conclusion}
In this paper, we introduce SDI-Paste as a novel synthetic data augmentation pipeline for VIS. SDI-Paste combines a generative text-to-video model and a self-supervised object segmentor in a carefully designed pipeline that yields dynamic object instances. We copy and paste these instances across a base dataset to achieve strong improvement over baseline and other existing augmentation strategies. The individual modules of our pipeline can be swapped for better and newer modules as T2V generators and object segmentors improve over time. The essence of our framework is infinitely scaleable and adaptable which we hope will make SDI-Paste a beneficial data augmentation regimen for other dense video tasks. 

\clearpage  

%
%
{\small
\bibliographystyle{ieee_fullname}
\bibliography{svi_paste}
}
\end{document}